%% file: main.tex
\documentclass[10pt,twocolumn,letterpaper]{article}

\usepackage[pagenumbers]{cvpr} 
\usepackage{booktabs}     
\usepackage{tabularx}     
\usepackage{makecell}     
\usepackage{multirow}     
\usepackage{stfloats}     
\usepackage{array}
\usepackage{graphicx}

\newcolumntype{Y}{>{\centering\arraybackslash}X}         
\newcolumntype{R}{>{\raggedleft\arraybackslash}p{1.05cm}}
\input{preamble}
\definecolor{cvprblue}{rgb}{0.21,0.49,0.74}
\usepackage[pagebackref,breaklinks,colorlinks,allcolors=cvprblue]{hyperref}

\def\paperID{*****} 
\def\confName{CVPR}
\def\confYear{2026}

\title{Learning to Accelerate Vision-Language-Action Models through Adaptive Visual Token Caching}

\usepackage{marvosym} 

\author{
    \makebox[\textwidth][c]{%
        Yujie Wei\textsuperscript{1*}\hspace{0.8em}
        Jiahan Fan\textsuperscript{2*}\hspace{0.8em}
        Jiyu Guo\textsuperscript{2}\hspace{0.8em}
        Ruichen Zhen\textsuperscript{3}\hspace{0.8em}
        Rui Shao\textsuperscript{2}
    }\\
    \makebox[\textwidth][c]{%
        Xiu Su\textsuperscript{4}\hspace{0.8em}
        Zeke Xie\textsuperscript{5}\hspace{0.8em}
        Hongxun Yao\textsuperscript{1}\hspace{0.8em}
        Shuo Yang\textsuperscript{2\Letter}
    }\\[2mm]
    \makebox[\textwidth][c]{%
        \textsuperscript{1}Harbin Institute of Technology \hspace{1.5em}
        \textsuperscript{2}Harbin Institute of Technology, Shenzhen
    }\\
    \makebox[\textwidth][c]{%
        \textsuperscript{3}Meituan Academy of Robotics Shenzhen, Meituan \hspace{1.2em}
        \textsuperscript{4}Central South University \hspace{1.2em}
        \textsuperscript{5}HKUST(GZ)
    }
}

\begin{document}
\maketitle
\begingroup
    \renewcommand\thefootnote{} 
    \begin{NoHyper} 
        \footnotetext{
            \noindent
            \begin{tabular}[t]{@{}l@{ }l} 
                * & Equal Contribution \\
                \textsuperscript{\Letter} & Corresponding Author: shuoyang@hit.edu.cn
            \end{tabular}
        }
    \end{NoHyper} 
\endgroup

\input{sec/0_abstract}    
\input{sec/1_intro}
\input{sec/2_related}
\input{sec/3_method_2.0}
\input{sec/4_experiments}
\input{sec/5_conclusion}
{
    \small
    \bibliographystyle{ieeenat_fullname}
    \bibliography{main}
}

\appendix
\input{sec/X_suppl}

\end{document}

%% file: sec/0_abstract.tex
\begin{abstract}
Vision-Language-Action (VLA) models have demonstrated remarkable generalization capabilities in robotic manipulation tasks, yet their substantial computational overhead remains a critical obstacle to real-world deployment. Improving inference efficiency is therefore essential for practical robotic applications. Existing acceleration methods often rely on heuristic or static strategies—such as rule-based token caching or pruning—that are decoupled from task objectives and fail to adapt to dynamic scene changes.
In this work, we reformulate inference acceleration as a learnable policy optimization problem and propose a novel framework that integrates a dynamic, task-aware decision-making process directly into the VLA model. At its core are two lightweight, cooperative modules: a \textit{Cached Token Selector}, which determines \textit{which} tokens should be reused, and a \textit{Cache Ratio Predictor}, which controls \textit{how many} tokens to reuse. Training these modules is non-trivial due to their discrete decisions. We address this by adopting a differentiable relaxation that allows gradient-based end-to-end optimization.
Extensive experiments on the LIBERO and SIMPLER benchmarks, as well as real-robot evaluations, show that our method achieves a 1.76$\times$ wall-clock inference speedup while simultaneously improving the average success rate by 1.9 percentage points (from 75.0\% to 76.9\%) on LIBERO and by 5.0 percentage points on real-world tasks, significantly outperforming existing baselines. This work highlights the potential of learning task-aware computational allocation policies, paving the way for VLA models that are both powerful and efficient.Code is available at \href{https://github.com/JiahanFan/LAC}{GitHub}.
\end{abstract}

%% file: sec/1_intro.tex
\section{Introduction}
\label{sec:intro}

\begin{figure}[t]
    \centering
    \includegraphics[width=0.93\columnwidth]{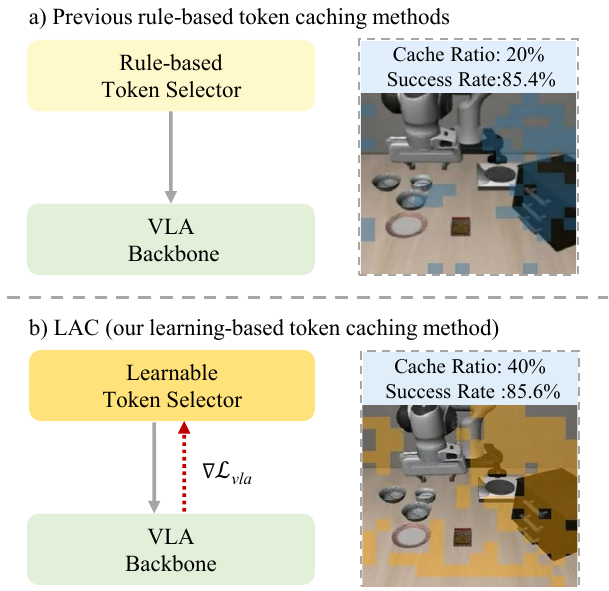}
    \caption{
Our learnable caching (LAC) outperforms rule-based baselines in both efficiency and performance. 
\textbf{(a)}~Rule-based selectors are decoupled from the task objective. They often rely on proxies, but \textit{where the model attends is not necessarily what the task requires}. This leads to suboptimal, task-agnostic caching (20\% cache ratio). 
\textbf{(b)}~Our selector is optimized with direct task gradients ($\nabla\mathcal{L}_{\text{VLA}}$), ensuring its decisions align with actual task requirements. This enables a more aggressive adaptive caching strategy, leading to a much higher cache ratio (40\%) and greater efficiency, while simultaneously improving the success rate.
    }
    \label{fig:one}
\end{figure}

Large Vision-Language-Action (VLA) models have recently shown remarkable progress in unifying perception, reasoning, and control for embodied intelligence~\cite{zitkovich2023rt,driess2023palme,kim2024openvla,li2024cogact,black2024pi_0,zhao2025cot,intelligence2025pi_,zhou2025chatvla,zhong2025dexgraspvla}. 
By coupling large-scale multimodal pretraining with autoregressive action decoding, models such as RT-2~\cite{zitkovich2023rt}, PaLM-E~\cite{driess2023palme}, and OpenVLA~\cite{kim2024openvla} enable robots to interpret open-ended language commands and perform complex manipulation tasks. 
However, deploying such models in real-world robotic systems remains challenging due to their high computational overhead, which conflicts with the low-latency requirements of real-time control, hindering widespread adoption.

A major inefficiency arises from redundant computation across consecutive frames, as VLA models reprocess the entire visual input at every step, even when most background regions remain static. 
Existing acceleration methods attempt to mitigate this by caching or pruning visual tokens, but they typically rely on fixed, rule-based heuristics~\cite{chen2024image,zhang2024sparsevlm,xu2025vla}. 
As illustrated in Figure~\ref{fig:one} (a), these rule-based selectors are fundamentally decoupled from the downstream task objective.
This means their decisions are based on proxies, such as attention scores or visual saliency, which often fail to capture true task-relevance. 
A policy might waste computation on visually distracting but functionally irrelevant regions, or erroneously discard subtle but critical cues for the task. 
In essence, \textit{\textbf{where the model attends is not necessarily what the task requires}}. 
This fundamental misalignment leads to a suboptimal trade-off between speed and accuracy, limiting both the potential efficiency gains and the overall reliability of the system.

We argue that the key to efficient VLA inference lies not in manually designing better heuristics, but in learning a dynamic computation policy that is directly optimized for task success. 
We hypothesize that an optimal policy should learn to allocate computational resources based on scene motion and task requirements, reusing redundant information while recomputing only the most salient parts.

Based on this insight, we propose \textbf{LAC} (\textbf{L}earnable \textbf{A}daptive \textbf{C}aching for Vision-Language-Action models), a lightweight framework that transforms heuristic acceleration into a learnable, task-driven policy. 
As shown in Figure~\ref{fig:one} (b), LAC introduces a \textit{learnable token selector} that receives direct feedback from the final task loss ($\nabla\mathcal{L}_{\textit{\text{vla}}}$) which measures the ability of VLA models. 
This end-to-end optimization enables the selector to make more intelligent, task-aware decisions. 
As a result, LAC can adopt a more aggressive caching strategy (40\% cache ratio) that not only improves efficiency but also enhances task performance, achieving a higher success rate.
At its core, LAC features two cooperative modules: the \textit{Cached Token Selector}, which learns token-level dynamic saliency, and the \textit{Cache Ratio Predictor}, which determines the overall reuse budget.
To effectively assess scene dynamics, these modules leverage computationally lightweight optical flow. This approach provides a direct, pixel-level signal of motion, which is crucial for identifying which visual tokens correspond to moving objects like the robot's gripper or manipulated items. Crucially, we opt for this direct motion signal over language-based guidance, as the latter would likely necessitate a larger model and incur additional computational overhead, running counter to our goal of efficiency. 
Both are trained jointly through a task-driven policy learning framework, ensuring that every computation decision contributes directly to the final goal.

Before delving into the details, we summarize our contributions as follows:
\begin{itemize}
    \item We propose \textbf{LAC}, the first learnable adaptive caching framework that transforms heuristic acceleration into an end-to-end trainable policy for VLA models.
    \item We design two lightweight and cooperative decision modules—the \textit{Cached Token Selector}, which identifies which visual tokens should be recomputed or reused, and the \textit{Cache Ratio Predictor}, which determines how many tokens to reuse based on scene-level motion entropy. Together, they enable fine-grained and adaptive token reuse for efficient VLA inference.
    \item Extensive experiments on LIBERO, SIMPLER, and real-world robotic platforms demonstrate that LAC achieves up to a 1.76× wall-clock acceleration while simultaneously improving task performance, outperforming existing rule-based pruning and caching baselines.
\end{itemize}

%% file: sec/2_related.tex
\section{Related Work}
\label{sec:related}

\textbf{Vision-Language-Action Models.} Vision-Language-Action (VLA) models represent a significant advancement in robotics, building upon the multimodal reasoning capabilities of large-scale vision-language models (VLMs)~\cite{alayrac2022flamingo, li2023blip, liu2024improved, bai2023qwen,dai2023instructblip,Du2021GLMGL,Wang2024Qwen2VLEV,shen2024mome}. By introducing an action modality, these models~\cite{zitkovich2023rt, kim2024openvla, li2024cogact, team2024octo, li2023vision,liu2024rdt,li2025hamster,liu2025hybridvla,wang2025robobert,wen2023syreanet,shi2025hi,zhou2025chatvla} are able to perform end-to-end visuomotor control. The common paradigm involves adapting powerful VLM backbones, such as Llama~\cite{touvron2023llama, grattafiori2024llama}, through fine-tuning on extensive robotics datasets like Open X-Embodiment~\cite{o2024open}. To generate actions, these models employ diverse strategies, from discretizing robot movements into language-like tokens~\cite{zitkovich2023rt, kim2024openvla} to integrating specialized decoders like diffusion action transformers~\cite{li2024cogact,wen2025diffusionvla,wen2025dexvla} or flow models~\cite{black2024pi_0,intelligence2025pi_}. This has led to successes in complex manipulation tasks~\cite{huang2024rekep, zhao2023learning}, but the substantial computational demands of VLA models remain a critical hurdle for practical, real-time deployment in resource-constrained settings.

\noindent\textbf{Acceleration of VLA Models.} Prior work on accelerating VLMs often falls short for robotic applications. Strategies like token pruning~\cite{chen2024image, zhang2024sparsevlm, cao2024madtp, lin2024mope,xing2024pyramiddrop} and merging~\cite{cao2023pumer,hu2025mplug,li2024llama,shang2025llava} are ill-suited for robotics. They fail to exploit critical temporal redundancy across video frames. Moreover, their pruning logic can harm the spatial fidelity required for manipulation and introduces computational overhead that often outweighs the benefits for short robotic action sequences. VLA-specific approaches, including architectural optimizations~\cite{liu2024robomamba, wen2025tinyvla}, dynamic scheduling~\cite{yue2024deer, zhang2025mole,li2025cogvla}, and quantization~\cite{park2024quantization}, demand costly retraining or rely on heuristic-based rules. High-frequency control methods~\cite{pertsch2025fast, zhang2024hirt, kim2025fine} also leave the vision processing bottleneck largely unaddressed. The most related approach, VLA-Cache~\cite{xu2025vla}, introduces key-value caching across timesteps but employs a static, rule-based policy that cannot effectively adapt to changing scene dynamics.

In contrast, our proposed LAC transforms acceleration into a learnable, task-oriented policy, using jointly optimized modules to create a fine-grained, adaptive caching strategy that dynamically and effectively balances computation and accuracy.

%% file: sec/3_method_2.0.tex
\section{Method}
\label{sec:method}
\begin{figure*}[t]
    \centering
    \includegraphics[width=1\textwidth]{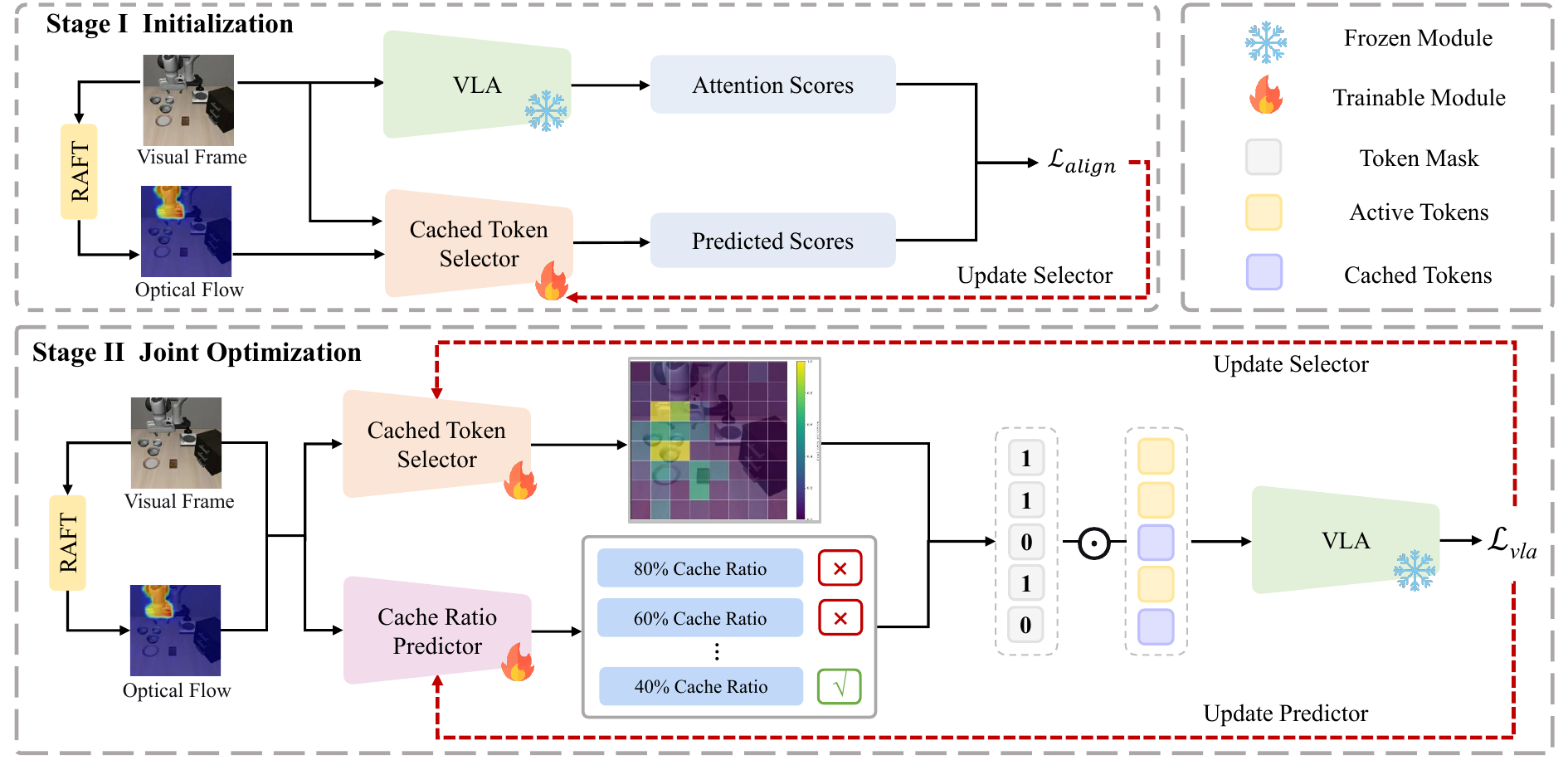}
    \caption{
\textbf{Overview of the two-stage training framework for LAC.}
\textbf{Stage I: Initialization.}
The Cached Token Selector is initially pretrained to align with the VLA's internal attention map, providing a stable and informative initialization.
\textbf{Stage II: Joint Optimization.}
Following this, the Cached Token Selector and Cache Ratio Predictor are jointly optimized with the VLA backbone held frozen.
The selector learns token-level saliency for dynamic reuse, while the predictor estimates the optimal cache ratio from a discrete set.
Gradients from the task loss $\mathcal{L}_{\text{VLA}}$ are backpropagated through both modules via the differentiable selection mechanism, enabling end-to-end policy learning for adaptive computation.
}
    \label{fig:framework} 
\end{figure*}
To address the high computational cost faced by Vision-Language-Action (VLA) models in real-time decision-making, we propose a dynamic computation optimization framework that enables adaptive efficiency within the perception–action loop. For embodied agents, efficient perception is the foundation of low-latency and responsive control. Our approach accelerates inference by intelligently reusing redundant visual information rather than performing uniform, dense computation for every incoming frame. The core of the method lies in two lightweight, cooperative modules that dynamically determine token-wise processing strategies based on scene content. The following sections detail the overall framework, key components, and the two-stage training procedure.

\subsection{Overall Framework}
\label{sec:framework_overview}

To enhance the inference efficiency of VLA models, we propose a learnable adaptive caching framework that dynamically decides which visual tokens to recompute and which to reuse from a cache. This adaptive computation strategy aims to reduce computational overhead while simultaneously improving task performance. A high-level overview of our proposed method is illustrated in Figure~\ref{fig:framework}.

At the heart of our framework are two lightweight, cooperative modules: the \emph{Cached Token Selector} and the \emph{Cache Ratio Predictor}. The Cached Token Selector is responsible for generating token-level saliency scores to identify which visual information is critical and requires recomputation. Concurrently, the Cache Ratio Predictor assesses the overall scene dynamics to determine an appropriate proportion of tokens to cache. These two modules operate as a plug-and-play addition, guiding the computational flow for a pre-trained and frozen VLA backbone, thereby preserving its powerful learned representations.

To train these decision-making modules effectively, we devise a two-stage training procedure. In the first stage, \textit{Initialization}, we train the Cached Token Selector to mimic the attention distribution of the expert VLA model. This provides the selector with a strong initial policy for identifying salient regions. In the second stage, \textit{Joint Optimization}, we fine-tune the selector and train the Cache Ratio Predictor end-to-end. The optimization objective in this stage is the final task performance, ensuring that the learned caching policy is directly aligned with the ultimate goal.

\subsection{Preliminaries: KV Caching for Efficiency}

In autoregressive Transformer models, Key-Value (KV) caching is a standard technique to accelerate inference by reusing key ($\mathbf{K}$) and value ($\mathbf{V}$) representations across decoding steps. Given an input sequence $\mathbf{X} = [x_1, x_2, \dots, x_T]$, the self-attention mechanism computes
\begin{gather}
\mathbf{Q} = \mathbf{X} W_Q, \quad 
\mathbf{K} = \mathbf{X} W_K, \quad 
\mathbf{V} = \mathbf{X} W_V, \\
\mathrm{Attention}(\mathbf{Q}, \mathbf{K}, \mathbf{V}) =
\mathrm{Softmax}\!\left(\frac{\mathbf{Q}\mathbf{K}^\top}{\sqrt{d}}\right)\mathbf{V}.
\end{gather}

During inference, only the new token's $\mathbf{k}_{\text{new}}$ and $\mathbf{v}_{\text{new}}$ are computed and appended to the cache as follows:
\begin{equation}
\mathbf{K}_t = [\mathbf{K}_{t-1}, \mathbf{k}_{\text{new}}], \quad
\mathbf{V}_t = [\mathbf{V}_{t-1}, \mathbf{v}_{\text{new}}].
\end{equation}
This incremental update avoids recomputation of historical states, significantly improving decoding efficiency.

While KV caching works well for language decoding within a single context window, it remains limited for Vision-Language-Action (VLA) models. In robotic control, consecutive frames are highly redundant, yet current VLA models typically re-encode full visual inputs at every step, wasting computation and increasing latency. 

This inefficiency motivates our work: most static regions need not be recomputed each timestep. By selectively updating only salient visual tokens and reusing cached representations for redundant ones, the model can achieve efficiency gains without sacrificing perception or control accuracy. Building on this intuition, our method learns an adaptive caching policy that dynamically decides \emph{which} and \emph{how many} to recompute, as detailed in the following sections.

\subsection{Dynamic Token Caching Modules}
\label{sec:modules}

At the core of our framework are two lightweight, cooperative modules: the \emph{Cached Token Selector} and the \emph{Cache Ratio Predictor}. To capture scene dynamics, they leverage a motion-aware representation $V_t = [I_t; O_t]$, formed by concatenating the visual frame $I_t$ and the optical flow $O_t$. We compute $O_t$ with a lightweight \mbox{RAFT-small}~\cite{teed2020raft} model, as it provides a direct and efficient motion signal, avoiding the higher computational overhead associated with using language for guidance. These modules jointly decide \textit{which} visual tokens to recompute and \textit{how many} to reuse from the cache. For an input token sequence $X_t = \{x_t^{(1)}, \dots, x_t^{(N)}\}$, they operate on $V_t$ to produce a binary mask $M_t \in \{0, 1\}^N$ that guides the per-token computation.

\subsubsection{Cached Token Selector}

The Cached Token Selector is a function $f_{\text{sel}}$ with parameters $\theta_{\text{sel}}$, responsible for assessing token-level importance. Given the motion-aware input $V_t$, it generates a vector of saliency scores $S_t$:
\begin{equation}
    S_t = f_{\text{sel}}(V_t; \theta_{\text{sel}}),
\end{equation}
where $S_t = \{s_t^{(1)}, \dots, s_t^{(N)}\}$ and $s_t^{(i)}$ is a continuous value in the range $[0, 1]$. A higher score $s_t^{(i)}$ indicates that token $x_t^{(i)}$'s information is critical for the current timestep---perhaps due to manipulator movement or object interaction---and should accordingly be recomputed. Conversely, tokens with lower scores are identified as candidates for caching. To ensure minimal computational overhead, $f_{\text{sel}}$ is implemented as a small and efficient convolutional neural network (CNN).

\subsubsection{Cache Ratio Predictor}

While the selector provides a relative ranking, the Cache Ratio Predictor determines the overall computational budget. This module, $f_{\text{pred}}(\cdot; \theta_{\text{pred}})$, also assesses global scene dynamics using the input $V_t$ to select an appropriate cache ratio from a discrete set $\mathcal{R} = \{r_1, \dots, r_C\}$ by outputing a vector of logits $L_t$ over the set $\mathcal{R}$:
\begin{equation}
    L_t = f_{\text{pred}}(V_t; \theta_{\text{pred}}),
\end{equation}
where $L_t = \{l_t^{(1)}, \dots, l_t^{(C)}\}$. Each logit $l_t^{(j)}$ corresponds to the model's confidence in applying cache ratio $r_j$. During inference, the ratio with the highest logit is selected. In a static scene, the predictor is trained to select a high ratio to maximize efficiency, while in a dynamic scene, it learns to select a lower ratio to maintain high task performance.

\subsection{Two-Stage Training Procedure}
\label{sec:training}

Learning a discrete caching policy from scratch is unstable under sparse supervision. We address this cold-start problem with a two-stage training paradigm: we first initialize the policy by distilling expert knowledge, then fine-tune it on task objectives while keeping the pretrained VLA frozen.

\noindent{\textbf{Stage I: Initialization via Attention Alignment.}}
This stage provides the \emph{Cached Token Selector} with an initialization by distilling knowledge from the expert VLA. Although the VLA’s attention scores are not optimal policies, they offer a useful proxy for general visual saliency. As shown in Figure~\ref{fig:framework}, the selector $f_{\text{sel}}$ is trained to mimic the VLA’s attention map $S_{\text{VLA}}$ by minimizing a MSE loss:
\begin{equation}
    \mathcal{L}_{\text{align}} = \text{MSE}(f_{\text{sel}}(V_t; \theta_{\text{sel}}), S_{\text{VLA}}).
\end{equation}
This warm-up provides the selector with a sensible initial policy, which is crucial for stabilizing the subsequent end-to-end optimization process.

\noindent{\textbf{Stage II: Joint Optimization for Task Performance.}}
After initialization, both modules are jointly optimized to balance task accuracy and computational efficiency. To make discrete selections differentiable, we employ the Gumbel-Softmax trick with straight-through estimation (see Sec.~\ref{sec:differentiable_selection}). The total loss combines the task loss $\mathcal{L}_{\text{VLA}}$ with a regularization term encouraging higher cache ratios:
\begin{equation}
    \mathcal{L}_{\text{total}} = \mathcal{L}_{\text{VLA}} + \lambda \mathcal{L}_{\text{ratio}},
\end{equation}
where
\begin{equation}
    \mathcal{L}_{\text{ratio}} = - \mathbb{E}_{\tilde{p}_t}[r] = - \sum_{j=1}^{C} \tilde{p}_t^{(j)} r_j.
\end{equation}
Here, $\tilde{p}_t$ denotes the soft probabilities from Gumbel-Softmax and $r_j$ the candidate cache ratios. The weight $\lambda$ controls the efficiency–performance trade-off. Gradients are backpropagated to update both modules.
This end-to-end training aligns the learned caching policy directly with the downstream control objective.

\subsection{Differentiable Discrete Decision Learning}
\label{sec:differentiable_selection}

A key challenge is that the decisions made by our modules are inherently discrete. The \emph{Cache Ratio Predictor} uses an $\arg\max$ operation to select a ratio from a discrete set $\mathcal{R}$, while the \emph{Cached Token Selector} relies on a non-differentiable \texttt{top-k} operation to identify tokens for caching. These operations obstruct gradient flow from the final task loss, preventing end-to-end optimization.

To overcome this obstacle, we introduce a differentiable policy optimization framework based on the Gumbel-Softmax trick~\cite{jang2016categorical} combined with the Straight-Through Estimator (STE)~\cite{bengio2013estimating}. This creates a hybrid ``hard forward, soft backward" strategy: it uses deterministic discrete selections in the forward pass for efficiency, while simultaneously enabling effective gradient flow through a continuous, ``soft" approximation during backpropagation.

For the \emph{Cache Ratio Predictor}, we use the Gumbel-Softmax trick. Given the module's output logits $l_t$, we then generate ``soft" probability vector $\tilde{p}_t$ by adding Gumbel noise with a temperature $\tau$:
\begin{equation}
\tilde{p}_t^{(j)} = \frac{\exp((l_t^{(j)} + g_j) / \tau)}{\sum_{k=1}^{C} \exp((l_t^{(k)} + g_k) / \tau)},
\end{equation}
where $g_j \sim \text{Gumbel}(0, 1)$. Then, in the forward pass, we perform a hard $\arg\max$ to get a one-hot vector $p_t$ for deterministic ratio selection:
\begin{equation}
p_t = \text{one\_hot}(\arg\max_j(\tilde{p}_t^{(j)})).
\end{equation}
During backpropagation, the Straight-Through Estimator (STE) passes gradients directly to the soft probabilities $\tilde{p}_t$.

For the \emph{Cached Token Selector}, we apply a similar principle. While the forward pass uses a hard mask $\mathbf{M}_t$ from the `top-k' operation, the backward pass constructs a differentiable ``soft" mask $\tilde{\mathbf{M}}_t$ using a steep sigmoid function:
\begin{equation}
\tilde{M}_t^{(i)} = \sigma\left(\frac{s_t^{(i)} - \theta_k}{\tau_s}\right),
\end{equation}
where $\theta_k$ is the score threshold for the $k_t$-th token and $\tau_s$ is a temperature parameter. Gradients are computed with respect to this soft mask. This hybrid strategy enables our decision modules to learn \textit{what} and \textit{how much} to cache in a unified, task-oriented framework.

\subsection{Inference Procedure}
\label{sec:inference}

Once the training process is complete, our decision modules are integrated with the frozen VLA backbone to create a highly efficient and stable inference pipeline.

\noindent{\textbf{Deterministic Decision-Making.}} For stable and consistent inference, we replace the stochastic sampling from training with deterministic $\arg\max$ operations. The \emph{Cache Ratio Predictor} selects the cache ratio $r_t$ corresponding to the highest logit. Subsequently, the \emph{Cached Token Selector} generates a binary mask $\mathbf{M}_t$ by selecting the $k_t = N \cdot r_t$ tokens with the lowest importance scores.

\noindent{\textbf{Efficient Forward Pass.}} The generated mask $\mathbf{M}_t$ partitions visual tokens to enable an efficient forward pass. \textit{Active tokens} are re-encoded to compute new Key (K) and Value (V) states, while the KV states of \textit{cached tokens} are directly reused from the previous timestep. This merged set of KV states is then fed to the action decoder, significantly reducing redundant computation.

\noindent{\textbf{Stochastic Recovery Mechanism.}} To enhance long-horizon robustness and mitigate error accumulation from persistent caching, we employ a stochastic recovery mechanism. At each step, with a small probability $p_{\text{recover}}$, a fraction of ``cached" tokens are randomly selected and forced to be recomputed. This low-cost refresh improves model stability without significantly impacting efficiency.

%% file: sec/4_experiments.tex
\section{Experiments}
\label{sec:experiments}

To comprehensively assess the performance and efficiency of LAC, we conduct a rigorous evaluation across both diverse simulated environments and a physical robotic platform. Our framework is flexibly integrated with two prominent open-source Vision-Language-Action (VLA) models: OpenVLA~\cite{kim2024openvla} and CogAct~\cite{li2024cogact}. These models are subsequently benchmarked on the LIBERO~\cite{liu2024libero} and SIMPLER~\cite{li2024evaluating} environments, respectively.

\subsection{Experimental Setup}

\noindent{\textbf{Compared Methods.}}
Given the shared architectural foundations of VLAs and Vision-Language Models (VLMs), acceleration techniques developed for VLMs can often be adapted for VLA inference. Consequently, we apply two leading token-level acceleration techniques, SparseVLM~\cite{zhang2024sparsevlm} and FastV~\cite{chen2024image}, to OpenVLA to serve as strong baseline comparisons on the LIBERO benchmark.

\noindent{\textbf{Evaluation Metrics.}}
We quantify the performance of our method using three key metrics: success rate, FLOPs, and CUDA time. Success rate measures the efficacy of task completion. On the efficiency front, FLOPs represent the theoretical computational cost, whereas CUDA time provides a practical measure of wall-clock runtime on the GPU. This combination of metrics is standard in the evaluation of VLM/VLA acceleration frameworks.

\subsection{Evaluation Benchmarks}

\noindent{\textbf{LIBERO.}}
The LIBERO benchmark~\cite{liu2024libero} is structured into four specialized suites: Spatial, Object, Goal, and Long, each specifically designed to probe a distinct dimension of manipulation generalization. To maintain consistency and ensure reproducibility, we adopt the standard setup of OpenVLA~\cite{kim2024openvla}, employing its officially provided model weights. Each suite consists of ten subtasks, and final performance is averaged over multiple trials.

\noindent{\textbf{SIMPLER.}}
The SIMPLER simulator~\cite{li2024evaluating} features two primary configurations: Visual Matching and Variant Aggregation. In line with the methodology of CogAct~\cite{li2024cogact}, we evaluate our method on a Google robot arm across four distinct manipulation tasks within both settings. By using CogAct as our base model, we test the generalizability of LAC to VLA models with diverse action decoding heads and under varied simulation conditions.

\noindent{\textbf{Real-world Robot Setting}}
We extend our evaluation to a physical Franka manipulator to assess performance under real-world conditions, including perception noise and actuation latency. For these experiments, we fine-tune the OpenVLA base model with LoRA~\cite{hu2022lora}. The evaluation is conducted on four manipulation tasks: KnockCrisp, PickMango, CoverBanana, and KnockBottle.

\begin{table*}[t]
\centering
\footnotesize
\setlength{\tabcolsep}{5pt}
\renewcommand{\arraystretch}{1.08}
\caption{    
\textbf{Results on the LIBERO benchmark.}
    Our method (LAC) simultaneously improves both performance and efficiency over all baselines. 
    Compared to the strong OpenVLA baseline, LAC increases the average success rate by 1.9 percentage points (from 75.0\% to 76.9\%) while achieving a 1.76$\times$ wall-clock speedup and reducing FLOPs by 25.3\%. }
\label{tab:libero_main}
\begin{tabularx}{\textwidth}{
@{}l
>{\centering\arraybackslash}X
>{\centering\arraybackslash}X
>{\centering\arraybackslash}X
>{\centering\arraybackslash}X
>{\centering\arraybackslash}X
>{\centering\arraybackslash}X
>{\centering\arraybackslash}X
@{}
}
\toprule
\textbf{Method} &
\makecell{\textbf{LIBERO}\\\textbf{Spatial}} &
\makecell{\textbf{LIBERO}\\\textbf{Object}} &
\makecell{\textbf{LIBERO}\\\textbf{Goal}} &
\makecell{\textbf{LIBERO}\\\textbf{Long}} &
\textbf{Average} &
\textbf{FLOPs(T)} $\downarrow$ &
\makecell{\textbf{CUDA}\\\textbf{Time (ms)} $\downarrow$} \\
\midrule
Baseline (OpenVLA)                & 84.4\% & \textbf{86.6\%} & 75.6\% & 53.2\% & 75.0\% & 1.864 & 51.91 \\
SparseVLM~\cite{zhang2024sparsevlm} & 79.8\% & 67.0\% & 72.6\% & 39.4\% & 64.7\% & 1.407 & 83.39 \\
FastV~\cite{chen2024image}        & 83.4\% & 84.0\% & 74.2\% & 51.6\% & 73.3\% & 1.864 & 53.28 \\
VLA-Cache~\cite{xu2025vla}        & 83.8\% & 85.8\% & 76.4\% & 52.8\% & 74.7\% & \textbf{1.355} & 31.83 \\
\midrule
\textbf{Ours (LAC)}            & \textbf{85.6\%} & 86.2\% & \textbf{76.6\%} & \textbf{59.2\%} & \textbf{76.9\%} & 1.392 & \textbf{29.51} \\
\bottomrule
\end{tabularx}
\end{table*}

\subsection{Results on Simulation Environments}

\noindent{\textbf{Main Results on LIBERO.}}
Table~\ref{tab:libero_main} summarizes our main results on the LIBERO benchmark. Our method achieves a 1.76× wall-clock speedup and improves the average success rate by 1.9 percentage points. Notably, baselines such as SparseVLM show increased latency despite lower theoretical FLOPs. Their pruning logic introduces overhead that outweighs the benefits for the short action sequences in robotics, and can also harm the spatial fidelity required for manipulation tasks. These results confirm that LAC delivers substantial acceleration for VLA models while simultaneously enhancing task performance and generalization.

\begin{figure*}[t]
    \centering
    \includegraphics[width=\linewidth]{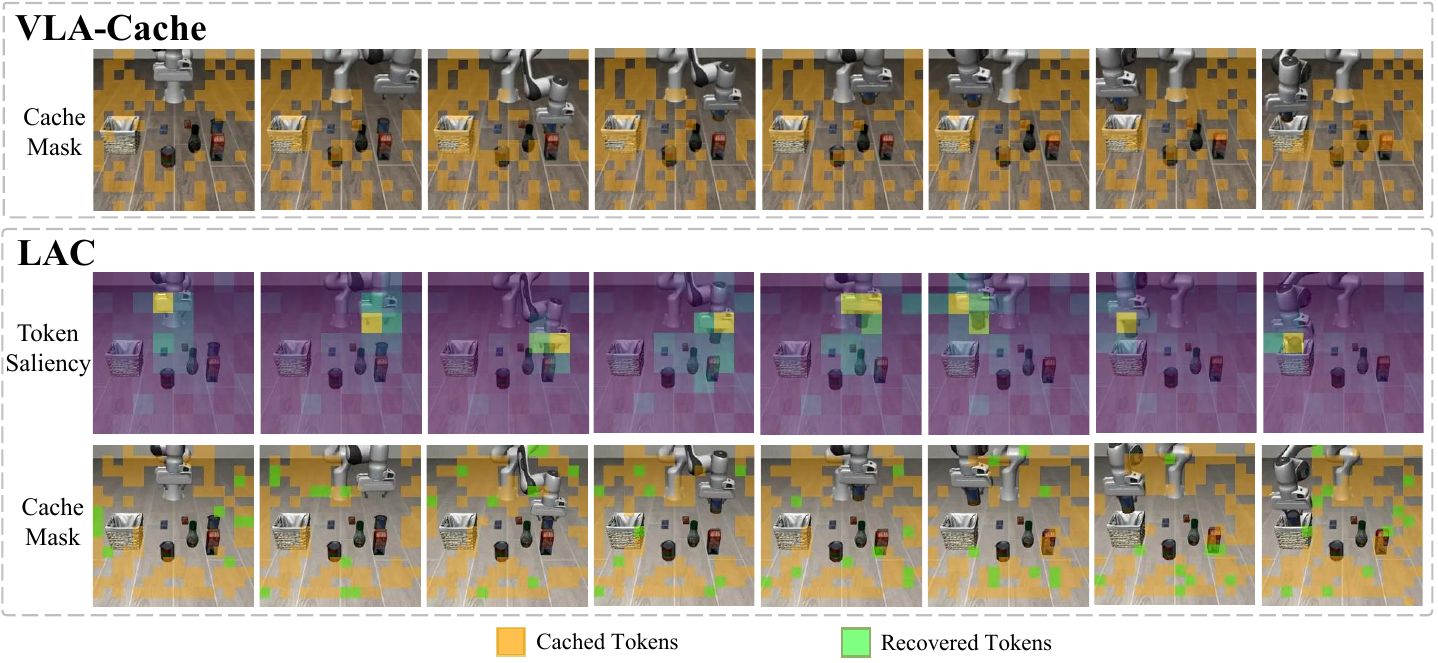}
    \caption{
    \textbf{Qualitative visualization of learned vs. rule-based caching.}
    LAC's policy (bottom) uses a learned saliency map to focus computation on the moving gripper while caching the static background (orange).
    In contrast, the rule-based method (top) detrimentally caches the static target basket, a task-agnostic error that could lead to task failure.
    This demonstrates the superiority of LAC's learned approach, which is further stabilized by recovered tokens (green).
    }
    \label{fig:libero_vis}
\end{figure*}

\noindent{\textbf{Qualitative Results on LIBERO.}}
Figure~\ref{fig:libero_vis} highlights the superiority of LAC's learned computational policy. LAC's saliency map (middle row) correctly identifies the moving gripper as the most critical region, ensuring it is always recomputed while the static background is efficiently cached (orange).
Conversely, the rule-based method (top row) makes a critical error: its task-agnostic heuristic caches the target basket simply because it is initially static. This decision deprives the model of essential visual updates for the upcoming interaction, causing the policy to fail as the robot arm gets stuck at the basket's rim, unable to find the correct placement. LAC's policy, guided by the task objective, avoids this failure, and its stability is further enhanced by stochastic recovered tokens (green). The figure clearly illustrates that a learned policy is essential for selectively recomputing salient information, unlike rigid heuristics that can discard vital context.

\begin{figure}[t]
    \centering
    \includegraphics[width=\linewidth]{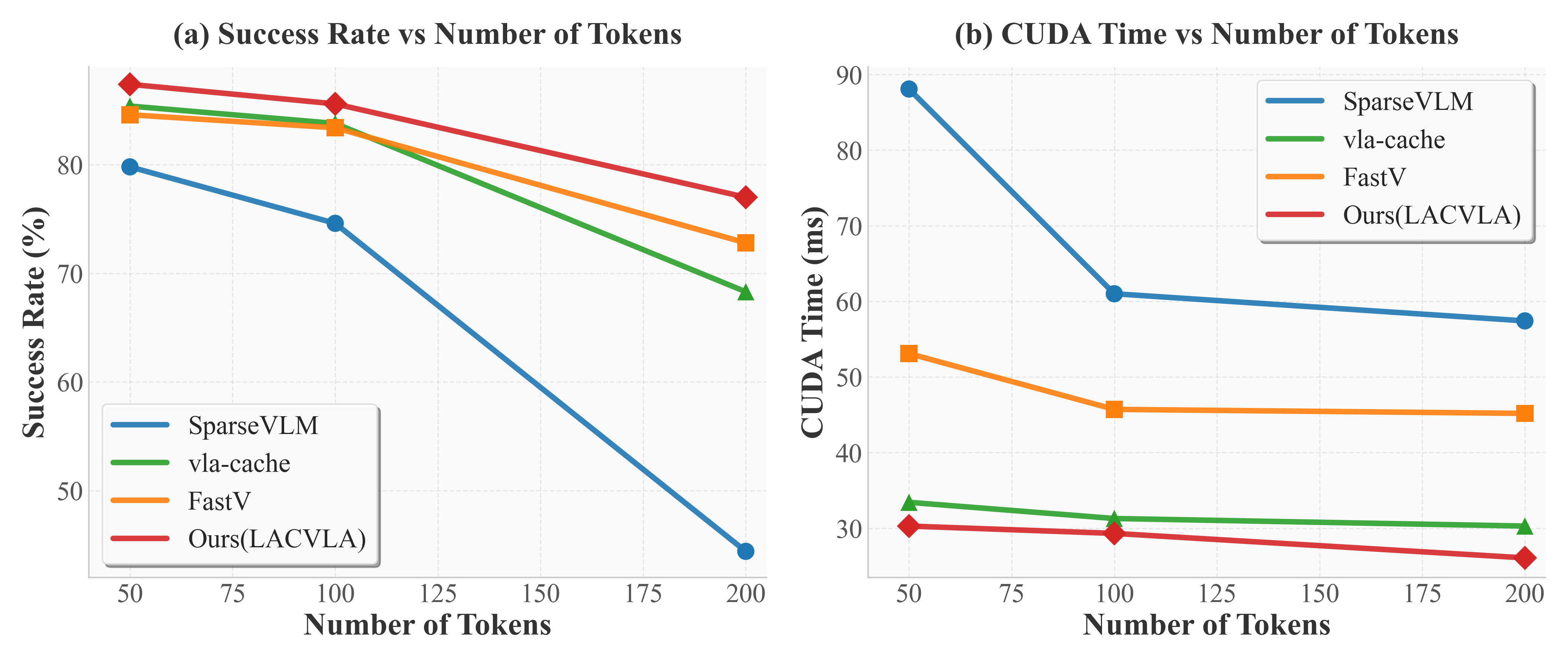}
    \caption{
        \textbf{Effect of Token Reusing/Pruning Ratio on LIBERO-Spatial.}
        Comparison of different methods under varying token reusing (or pruning) ratios. 
        \textbf{Left:} Task success rate. 
        \textbf{Right:} CUDA inference time. 
        Our LAC maintains consistently high performance as the reuse ratio increases, 
        while pruning-based methods such as SparseVLM degrade rapidly. 
        In addition, LAC achieves the lowest CUDA latency across all settings, 
        confirming its genuine wall-clock acceleration.
    }
    \label{fig:libero_curve}
\end{figure}

\noindent{\textbf{Effect of Token Reusing Ratio.}}
Figure~\ref{fig:libero_curve} compares methods under different token reusing or pruning rates. 
As the reuse ratio increases, LAC maintains stable performance, whereas pruning-based methods (e.g., SparseVLM) exhibit steep drops in success rate. 
Moreover, our CUDA inference time remains consistently lower, indicating genuine wall-clock efficiency rather than theoretical FLOPs reduction.

\begin{table}[htbp]
\centering
\small
\caption{Ablation on key components. Each row incrementally adds one module.}
\label{tab:method_ablation}
\setlength{\tabcolsep}{5pt} 
\begin{tabular}{@{}l c c c@{}}
\toprule
\textbf{Method} & \textbf{Success (\%)} & \textbf{FLOPs(T)} $\downarrow$ & \textbf{Time (ms)} $\downarrow$ \\
\midrule
Selector only & 82.20 & 1.283 & 28.48 \\
+ Reuse Predictor & 83.40 & 1.325 & 29.04 \\
+ Recovery (Full) & \textbf{85.60} & 1.377 & 29.32 \\
\bottomrule
\end{tabular}
\end{table}

\noindent{\textbf{Ablation Studies.}}
We ablate LAC by progressively enabling its three core components (Table~\ref{tab:method_ablation}).
Starting with the Token Selector alone achieves a $82.2\%$ success rate, showing that learning token-level saliency already captures essential visuomotor cues for efficient reasoning.
Adding the Reuse Predictor improves performance to 83.4\%(+1.2points). The slight increase in FLOPs and time reflects its learned strategy: it adaptively reduces the cache ratio during critical moments to ensure task success. This demonstrates a successful trade-off, sacrificing a small amount of speed for a notable gain in robustness. This indicates that adaptive, scene-level reuse budgeting is effective.
Finally, introducing the stochastic recovery mechanism yields the full model performance of $85.6\%$ ($+2.2$ points over the previous variant), which enhances robustness and consistency across diverse task settings.
These results verify that each component contributes progressively to efficiency–accuracy balance, forming a cohesive and complementary acceleration framework.

\begin{table*}[htbp]
\caption{Results on the SIMPLER benchmark~\cite{li2024evaluating} using CogAct~\cite{li2024cogact} as the base model.
Best results in each block are \textbf{bolded}.}
\label{tab:simpler_eval}
\resizebox{\linewidth}{!}{
\begin{tabular}{@{}clccccccc@{}}
\toprule
\textbf{SIMPLER} & \textbf{Method} &
\textbf{PickCan} & \textbf{MoveNear} & \textbf{Drawer} & \textbf{DrawerApple} & \textbf{Average} &
\textbf{FLOPs(T)} $\downarrow$ &
\makecell{\textbf{CUDA}\\\textbf{Time (ms)} $\downarrow$} \\
\midrule
\multirow{3}{*}{\textbf{\makecell{Visual\\Matching}}} &
Baseline (CogAct) &
91.3\% & \textbf{85.0\%} & 71.8\% & 50.9\% & 74.8\% & 1.847 & 54.29 \\
& VLA-Cache &
92.0\% & 83.3\% & 70.5\% & 51.6\% & 74.4\% & \textbf{1.496} & 39.63 \\
& \textbf{Ours (LAC)} &
\textbf{92.3\%} & 84.2\% & \textbf{72.7\%} & \textbf{52.8\%} & \textbf{75.5\%} & 1.511 & \textbf{37.82} \\
\midrule
\multirow{3}{*}{\textbf{\makecell{Variant\\Aggregation}}} &
Baseline (CogAct) &
89.6\% & 80.8\% & 28.3\% & 46.6\% & 61.3\% & 1.807 & 53.54 \\
& VLA-Cache~ &
91.7\% & 79.3\% & \textbf{32.5\%} & 45.8\% & 62.3\% & \textbf{1.493} & 39.11 \\
& \textbf{Ours (LAC)} &
\textbf{92.1\%} & \textbf{81.5\%} & 31.2\% & \textbf{47.1\%} & \textbf{63.0\%} & 1.506 & \textbf{37.90} \\
\bottomrule
\end{tabular}
}
\end{table*}

\noindent{\textbf{Main Results on SIMPLER.}}
As shown in Table~\ref{tab:simpler_eval}, our method achieves comparable or even better success rates than CogAct while significantly reducing computational cost.  
Specifically, our approach achieves an average success rate of 75.5\% in the \emph{Visual Matching} setting (CogAct: 74.8\%) and 63.0\% in the more challenging \emph{Variant Aggregation} setting (CogAct: 61.3\%).  
Furthermore, the efficiency improvements are substantial: our method reduces FLOPs by 17.4\% (1.827 $\rightarrow$ 1.509) and achieves a 1.42$\times$ speedup in inference time (53.92\,ms $\rightarrow$ 37.86\,ms).  
These results highlight the strong portability of our approach, demonstrating its ability to serve as a general and architecture-agnostic acceleration strategy for VLA models with different action decoders (e.g., diffusion-based).

\begin{table*}[htbp]
\centering
\small
\setlength{\tabcolsep}{4pt}
\renewcommand{\arraystretch}{1.0}
\caption{Real-world robotic manipulation results on four tasks.}
\label{tab:real_robot}
\begin{tabular}{lccccccc}
\toprule
\textbf{Method}   
 & \textbf{KnockCrisp} & \textbf{PickMango} & \textbf{CoverBanana} & \textbf{KnockBottle} & \textbf{Average}  & \textbf{FLOPs(T)} $\downarrow$ 
 & \makecell{\textbf{CUDA}\\\textbf{Time (ms)} $\downarrow$} \\
\midrule
Baseline (OpenVLA) 
 & 48.0\% & \textbf{16.0\%} & \textbf{24.0\%} &44.0\% & 33.0\% 
 & 1.893 & 37.38 \\
 VLA-Cache
 & 48.0\% & 12.0\% & 20.0\% & 40.0\% & 30.0\%
 & \textbf{1.534} & 33.36 \\
\midrule
\textbf{Ours (LAC)} 
 & \textbf{52.0\%} & 12.0\% & \textbf{24.0\%} & \textbf{64.0\%} & \textbf{38.0\%}
 & 1.569 & \textbf{32.47} \\
\bottomrule
\end{tabular}
\end{table*}

\begin{figure*}[t!]
    \centering
    \includegraphics[width=\textwidth]{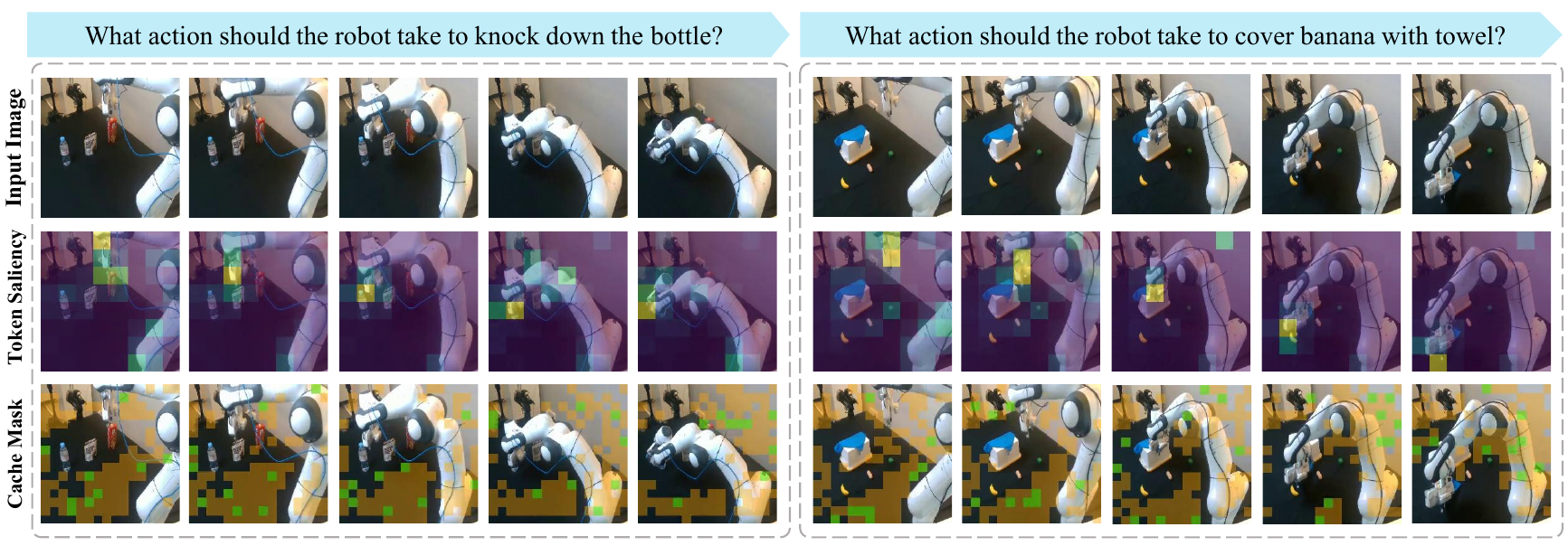}
    \caption{
        Qualitative visualization of our learned policy (LAC) on a real-world manipulation task. 
        The Token Saliency map (middle row) demonstrates that the model correctly identifies the moving end-effector and its immediate interaction space as the most critical regions. 
        Consequently, the Cache Mask (bottom row) shows an efficient policy where static background elements are cached, focusing computation only on task-relevant dynamics.
        Visualizations for two other tasks can be found in the Appendix.
    }
    \label{fig:real_world}
\end{figure*}

\subsection{Results on Real Robot}
We validate our approach on a physical robot across four manipulation tasks, with quantitative results detailed in Table~\ref{tab:real_robot}. Our method demonstrates superior performance, improving the average success rate by 5.0\% over the baseline while reducing FLOPs and inference latency.

To provide qualitative insight into these results, Figure~\ref{fig:real_world} visualizes the behavior of our learned policy. The Token Saliency map (middle row) confirms that our model learns to dynamically focus computation on critical, task-relevant regions—namely the robot’s end-effector and its immediate interaction space. Consequently, as shown in the Cache Mask (bottom row), this learned saliency translates directly into an efficient caching strategy: static background elements are reused, while only the salient, moving regions are recomputed. This ability to intelligently ignore visual distractions not only explains the computational gains but also contributes to greater policy robustness and higher task success rates in cluttered, real-world environments.

%% file: sec/5_conclusion.tex
\section{Conclusion}
\label{sec:conclusion}

This paper addresses the computational inefficiency of Vision-Language-Action (VLA) models for real-world robotics. We introduce a novel framework that transforms inference acceleration from a heuristic-driven process into a learnable, task-oriented policy. At its core, our method, LAC, uses two lightweight, jointly optimized modules to dynamically decide which visual tokens to recompute and which to reuse from a cache.

Extensive experiments on simulation benchmarks and a physical robot demonstrate our approach's effectiveness. Our method, LAC, achieves up to a 1.76× inference speedup while improving task success rates, outperforming static caching and pruning methods. These results validate our core hypothesis: by learning to adaptively reuse visual information, a learnable, task-driven policy can enhance the efficiency, robustness, and overall performance \mbox{of robotic control}.\nobreak

%% file: sec/X_suppl.tex
\clearpage
\section*{Appendix}
\section{Limitations}
While LAC enables efficient adaptive inference, it presents two main limitations. First, our reliance on optical flow as a motion prior means that extreme visual conditions may affect the precision of the token selector. Second, in scenarios with rapid global visual changes or high-speed camera ego-motion, the policy will intelligently opt to recompute most tokens to preserve accuracy; consequently, the efficiency gains in such highly dynamic settings will naturally diminish compared to more stable environments.

\section{Implementation Details}
\label{sec:appendix_inference}

This section further details the technical implementation of our Learnable Adaptive Caching (LAC) framework within the Transformer's forward pass. At each inference step \(t\), the binary mask \( \mathbf{M}_t \) generated by our policy dictates the selective computation process. This is realized through the following modifications to the VLA decoder's forward pass:

\begin{itemize}
    \item \textbf{Position Management and Attention Masking.}
    An internal state array tracks which tokens, as specified by \( \mathbf{M}_t \), require recomputation. Tokens designated for caching preserve their positional encodings from the previous step. This is crucial as it allows the attention masks to be dynamically pruned, restricting the scope of the self-attention mechanism to the reduced set of active tokens and thus saving computation.

    \item \textbf{Conditional Rotary Embedding.}
    To introduce fresh positional information for the current timestep, rotary embeddings are applied exclusively to the recomputed tokens. Cached tokens bypass this operation, retaining their previously encoded states and avoiding redundant processing.

    \item \textbf{Dynamic Key-Value (KV) Cache Updates.}
    The KV cache at each Transformer layer \(l\) is updated dynamically. Newly computed tokens update their respective entries with \(\{\mathbf{K}_t^l, \mathbf{V}_t^l\}\), while reused tokens inherit their values from the prior frame's cache, \(\{\mathbf{K}_{t-1}^l, \mathbf{V}_{t-1}^l\}\). This partial update strategy, which mixes new and old KV entries, yields valid attention outputs. This is enabled by the permutation-invariant nature of the Transformer architecture, which allows attention to be computed over any valid set of key-value pairs.
\end{itemize}

This entire mechanism is fully compatible with the standard KV caching used for autoregressive decoding. The most significant computational gain is realized during the generation of the initial action token at each timestep; subsequent tokens are then decoded autoregressively without incurring additional cost.

\section{Formal Efficiency Analysis}
\label{sec:formal_efficiency}

We provide a theoretical complexity analysis to quantify the computational benefits of LAC. Let $N$ denote the number of visual tokens per frame, $D$ the embedding dimension, $L$ the number of Transformer layers, and $M$ the intermediate dimension of the Feed-Forward Network (FFN).

\noindent
\textbf{Policy Overhead.} 
The overhead of our method stems from the fixed-cost optical flow estimation (RAFT) and the lightweight CNN selector. This cost scales linearly with the input image resolution $(H, W)$ and is independent of the Transformer depth or sequence length:
\begin{equation}
\mathcal{C}_{\text{policy}} \approx \mathcal{O}(H \cdot W \cdot C_{\text{cnn}}).
\end{equation}
Since this overhead is computed only once per frame and does not scale with the model depth $L$, it remains negligible compared to the backbone computation.

\noindent
\textbf{Baseline Computational Cost.} 
In a standard VLA Transformer layer, the computational cost consists of Multi-Head Self-Attention (MSA) and Feed-Forward Networks (FFN). For a sequence of $N$ tokens, the FLOPs per layer are:
\begin{equation}
\mathcal{C}_{\text{base}} \approx \underbrace{4ND^2 + 2N^2D}_{\text{MSA}} + \underbrace{2NDM}_{\text{FFN}}.
\end{equation}
Here, $4ND^2$ accounts for QKV and output projections, $2N^2D$ represents the attention score computation and aggregation, and $2NDM$ accounts for the two linear transformations in the FFN (projection $D \to M$ and $M \to D$).

\noindent
\textbf{LAC Computational Cost.} 
Our method processes only a subset of active tokens $N_{\text{act}} = (1-\rho)N$, where $\rho$ is the cache ratio.
\begin{itemize}
    \item \textbf{Projections \& FFN:} We compute linear projections and FFNs strictly for active tokens.
    \item \textbf{Attention:} Active queries ($N_{\text{act}}$) attend to the full context (active + cached tokens, total $N$).
\end{itemize}
The reduced cost per layer is:
\begin{equation}
\mathcal{C}_{\text{lac}} \approx \underbrace{4N_{\text{act}}D^2 + 2(N_{\text{act}} \cdot N)D}_{\text{Partial MSA}} + \underbrace{2N_{\text{act}}DM}_{\text{Partial FFN}}.
\end{equation}

\noindent
\textbf{Theoretical FLOPs Reduction.} 
The reduction in FLOPs per layer is derived by subtracting the LAC cost from the baseline:
\begin{equation}
\begin{aligned}
\Delta \text{FLOPs}_{\text{layer}} &= \mathcal{C}_{\text{base}} - \mathcal{C}_{\text{lac}}.
\end{aligned}
\end{equation}
Integrating the overhead, the total efficiency gain for the model is:
\begin{equation}
\Delta \text{FLOPs}_{\text{total}} \approx \sum_{l=1}^{L} \Delta \text{FLOPs}_{\text{layer}} - \mathcal{C}_{\text{policy}}.
\end{equation}
This analysis demonstrates that LAC significantly reduces the theoretical computational complexity, at the cost of a minimal, constant policy overhead.

\section{Additional Experiment Results}
\label{sec:add_exp}

\begin{figure*}[htbp]
    \centering
    \includegraphics[width=\textwidth]{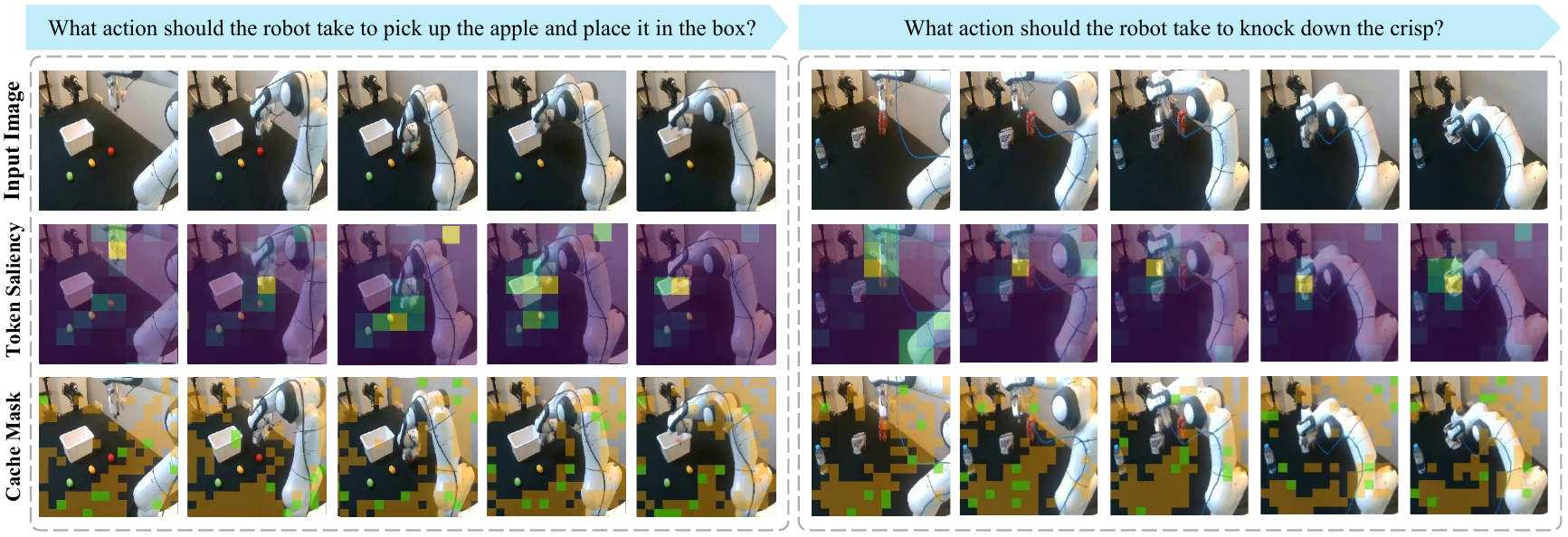}
    \caption{\textbf{Qualitative visualizations of the learned policy (LAC) on real-world tasks.} The examples show the policy correctly identifying task-relevant regions for recomputation. (Top) Input image sequence. (Middle) Learned Token Saliency map. (Bottom) Resulting Cache Mask, with \textcolor{orange}{orange} for cached tokens and \textcolor{green}{green} for stochastically recovered tokens.}
    \label{fig:appendix_realworld}
\end{figure*}

\subsection{Real-World Robot Tasks}
\label{sec:appendix_real_world}

Figure~\ref{fig:appendix_realworld} visualizes our policy's behavior on two real-world tasks: \textit{"pick up the apple and place it in the box"} (left) and \textit{"knock down the crisp"} (right). 
The learned Token Saliency (middle row) demonstrates the model's ability to effectively distinguish between important and unimportant visual regions. High saliency scores are concentrated on the robot's end-effector and the interaction targets, while the static background is assigned low importance. 
Consequently, the Cache Mask (bottom row) reveals an efficient strategy: unimportant visual regions are cached (marked in orange), while the critical, high-saliency regions remain unmasked for active recomputation. Additionally, green tiles appear within the cached regions, illustrating the stochastic recovery mechanism that randomly updates a fraction of tokens to ensure robustness.

\begin{figure*}[t]
    \centering
    \includegraphics[width=0.95\textwidth]{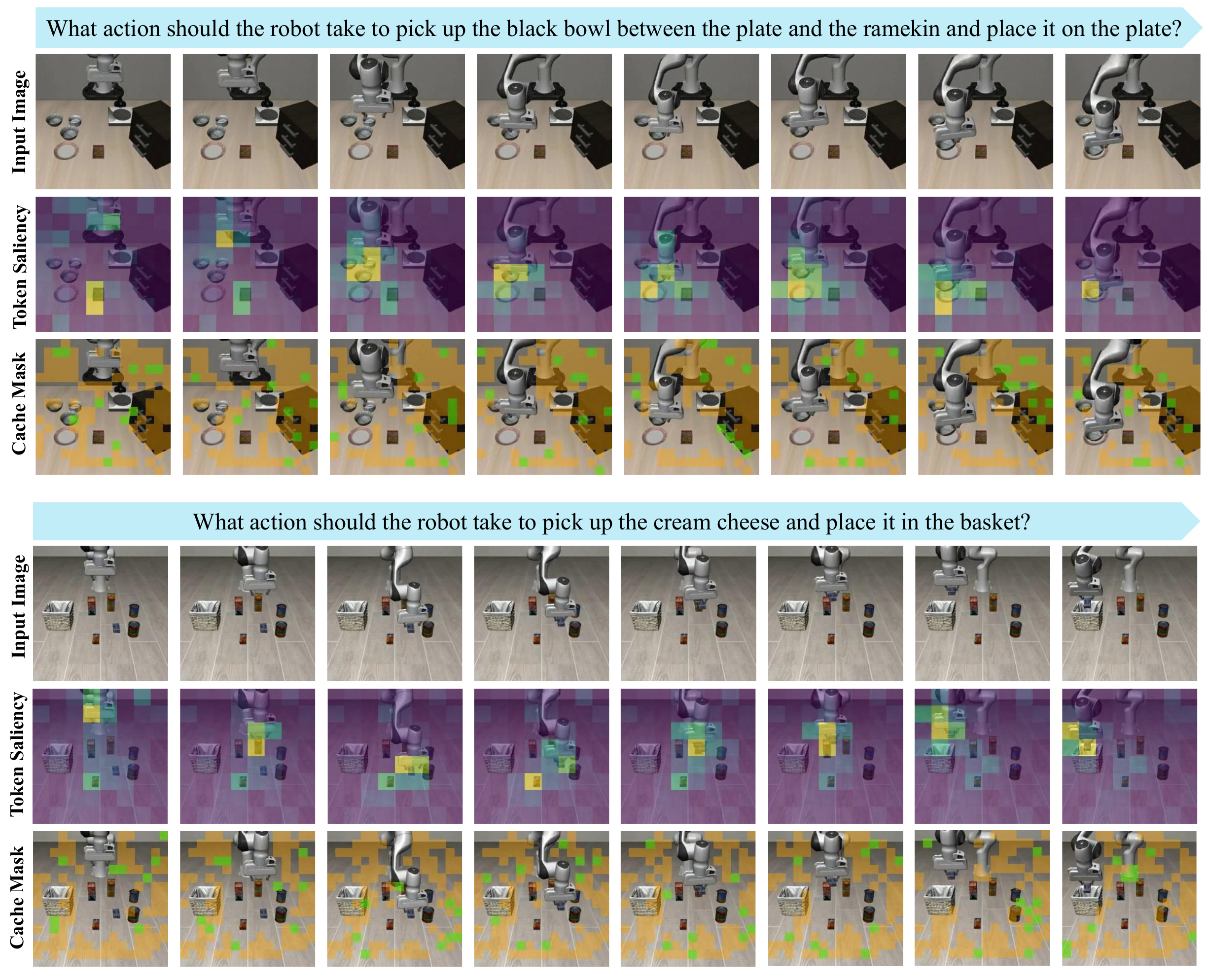}
    \caption{\textbf{Qualitative Results on LIBERO Benchmark (Part 1).} 
    We visualize the behavior of LAC on the first set of manipulation tasks. 
    (Top) Input image sequence; 
    (Middle) Learned Token Saliency map; 
    (Bottom) Generated Cache Mask (\textcolor{orange}{orange} for cached, \textcolor{green}{green} for stochastically recovered, and unmasked regions for active). 
    High saliency is consistently assigned to task-relevant regions like the gripper and target object, which are actively recomputed.}
    \label{fig:libero_part1}
\end{figure*}

\begin{figure*}[t]
    \centering
    \includegraphics[width=0.95\textwidth]{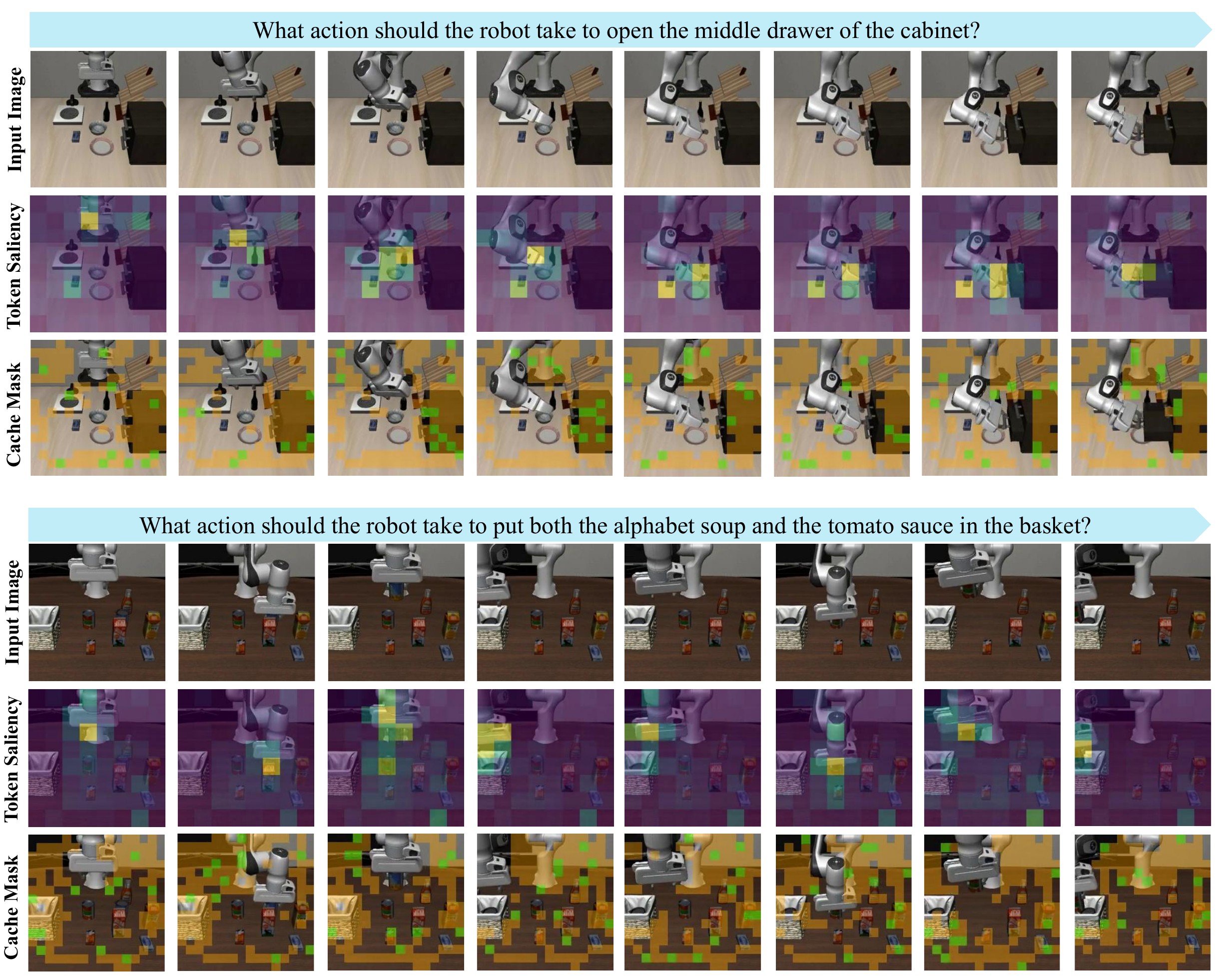}
    \caption{\textbf{Qualitative Results on LIBERO Benchmark (Part 2).} 
    Visualization on additional tasks. 
    Similar to Part 1, the policy successfully caches the unimportant visual regions (\textcolor{orange}{orange}) while actively re-computing the moving agents and interaction targets (shown as unmasked). \textcolor{green}{Green} tiles indicate tokens updated via the stochastic recovery mechanism, demonstrating the robustness of our adaptive caching strategy.}
    \label{fig:libero_part2}
\end{figure*}

\subsection{LIBERO Benchmark}
\label{sec:appendix_libero}

Figures~\ref{fig:libero_part1} and \ref{fig:libero_part2} present additional qualitative visualizations across four diverse tasks from the LIBERO benchmark. 
Consistent with our real-world observations, the learned policy accurately identifies task-relevant dynamics: high saliency scores are assigned to the robot gripper and target objects, while the static environment is cached for reuse. 
The resulting Cache Masks in both figures confirm that LAC efficiently allocates computation by re-computing critical regions (unmasked) and retrieving the static background from the cache (orange), with occasional updates provided by stochastic recovery (green) to prevent error accumulation.

\subsection{Additional Ablations and Comparisons}
\label{sec:supp_ablations}
To further justify our architectural decisions and training paradigm, we conduct additional ablation studies on the LIBERO benchmark. Specifically, we investigate the necessity of the Two-Stage Training strategy and the choice of input modality for the decision modules. The quantitative results are summarized in Table~\ref{tab:supp_ablation}.

\begin{table}[htbp]
\centering
\renewcommand{\arraystretch}{1.2} 
\small
\setlength{\tabcolsep}{10pt}
\begin{tabular}{l c}
\toprule
\textbf{Method Variant} & \textbf{Success Rate (\%)} \\
\midrule
w/o Stage I Initialization & 79.2 \\
Language-guided Policy   & 83.0 \\
\midrule
\textbf{Ours (LAC)}        & \textbf{85.6} \\
\bottomrule
\end{tabular}
\caption{\textbf{Additional Ablation Studies on LIBERO.} We compare our full method against a variant without Stage I initialization and a variant replacing optical flow with language guidance.}
\label{tab:supp_ablation}
\end{table}

\noindent\textbf{Impact of Initialization Strategy (Stage I).}
Our framework employs a two-stage training pipeline where the \textit{Cached Token Selector} is first initialized by distilling attention scores from the frozen VLA (Stage I) before end-to-end optimization. As shown in Table~\ref{tab:supp_ablation}, skipping this initialization step (``w/o Stage I Initialization'') results in a performance drop to 79.2\%. This confirms that learning a discrete selection policy from scratch under sparse supervision is unstable. The initialization phase provides a critical "visual saliency prior," ensuring the selector begins the joint optimization (Stage II) from a reasonable state rather than converging to suboptimal local minima.

\noindent\textbf{Motion vs. Language for Adaptive Caching.}
We also evaluated an alternative design where the LAC framework is conditioned on language instructions rather than optical flow. This variant achieves a success rate of 83.0\%, underperforming our motion-based approach. Consistent with the discussion in our main text, relying on language guidance presents a dilemma between accuracy and efficiency. Effective vision-language alignment typically requires large, computationally expensive encoders. In this ablation, to maintain an inference speed comparable to LAC, we employed a lightweight language conditioning module. However, this restricted capacity limits the model's ability to precisely align textual instructions with visual features. In contrast, optical flow provides a direct, pixel-level signal of motion that naturally captures task-relevant dynamics without the heavy computational overhead required for deep semantic processing.